\documentclass{article}
\usepackage{geometry}
\usepackage{cite}
\usepackage{algorithmic}
\usepackage{graphicx}
\usepackage{textcomp}
\usepackage{xcolor}
\usepackage{color,soul}
\usepackage{url}
\usepackage{subcaption}

\urldef{\myurl}\url{https://github.com/AloeUoB/RSSI_GAN}

\title{Transfer Learning of RSSI to Improve Indoor Localisation Performance}

\author{Thanaphon Suwannaphong, Ryan McConville and Ian Craddock \\ \\
 School of Engineering Mathematics and Technology\\
 University of Bristol, UK}

\date{} 

\begin{document}
\maketitle

\begin{abstract}
With the growing demand for health monitoring systems, in-home localisation is essential for tracking patient conditions. The unique spatial characteristics of each house required annotated data for Bluetooth Low Energy (BLE) Received Signal Strength Indicator (RSSI)-based monitoring system. However, collecting annotated training data is time-consuming, particularly for patients with limited health conditions. To address this, we propose Conditional Generative Adversarial Networks (ConGAN)-based augmentation, combined with our transfer learning framework (T-ConGAN), to enable the transfer of generic RSSI information between different homes, even when data is collected using different experimental protocols. This enhances the performance and scalability of such intelligent systems by reducing the need for annotation in each home. We are the first to demonstrate that BLE RSSI data can be shared across different homes, and that shared information can improve the indoor localisation performance.  Our T-ConGAN enhances the macro F1 score of room-level indoor localisation by up to 12.2\%, with a remarkable 51\% improvement in challenging areas such as stairways or outside spaces. This state-of-the-art RSSI augmentation model significantly enhances the robustness of in-home health monitoring systems.
\end{abstract}

\section*{Keywords}
  Indoor Localisation, IoT, RSSI Augmentation, Smart Home, Transfer Learning, GAN

\section{Introduction}
The increasing elderly population, often facing age-related chronic conditions such as Alzheimer's and Parkinson's diseases, presents healthcare challenges. In-home smart sensors with indoor localisation hold potential for improving elder care by utilizing wearable sensor systems to detect disease-related abnormal movement patterns in daily activities \cite{poyiadzi2020detecting}. This technology aids early intervention, promotes independent living and offers better health outcomes for the elderly.

Indoor localisation technology tracks user locations using received signal strength indicator (RSSI) measurements to estimate location based on signal strength between access points (APs) and a transmitter (e.g., phone or wearable), offering accurate and reliable indoor movement tracking. Among widely used radio signal frequencies, such as Bluetooth Low Energy (BLE), Wi-Fi, and Ultra-Wideband (UWB), BLE is the most cost-effective and practical for in-home health monitoring systems due to its low energy consumption, low cost, minimal maintenance, and ease of deployment. Although BLE does not offer accuracy as high as UWB and Wi-Fi CSI \cite{alarifi2016ultra}, its energy efficiency enables longer wearable use without frequent charging—a crucial factor for elderly users.  

Sufficient and extensive labelled data are required to create a radio map of the building for a highly accurate indoor localisation model. However, collecting labelled data in private residential building for healthcare settings faces difficulties as the process can be challenging and time-consuming for elderly patients \cite{bennett2017healthcare,van2020indoor}. For very detailed annotations the process may take up to an hour for a healthy individual \cite{byrne2018residential} in a typical home, which is a clear challenge for ill or elderly people who live independently. We note, however, that some studies have devised much shorter protocols for this process, but still involve a participant executing a set of steps correctly \cite{McConville2021}. Unlike other settings, such as large public building with more than 100 APs where healthy researchers can walkaround and collect training data much easier, our main focus is to benefit a quick and easy setup for in-home monitoring system with minimum effort for training data collection, which an elderly can simply perform independently. 

In a healthcare setting, not all areas within the home are equally important for health monitoring and emergency systems. Certain areas, such as stairs and outdoor spaces adjacent to the house, are particularly critical as they are high-risk zones that often require additional assistance. This is especially true for elderly individuals or those with mobility issues, making the collection of labelled data in these areas even more challenging for them. Therefore, our study also focus on improving the indoor localisation of these areas without additional labour from the patients.

Increasing the training data without collecting more labelled data is a practical solution to the challenges mentioned. This can be achieved through data augmentation techniques or by sharing data across other sources, such as different houses or datasets but both methods present their own challenges. Data augmentation is an efficient solution \cite{Suwannaphong2022}, but it's complex for time-series data like RSSI, requiring preservation of temporal relationship of the RSSI among different APs which represent the location information of the actual environment. Alternatively, sharing RSSI data from other sources seems simple, but the collected labelled RSSI data has never previously been shared among different houses due to the inherent characteristic of RSSI signals that predominantly corresponds to the reflection of physical elements, including walls and furniture, presented within a particular residence, making the signal unique for each house \cite{sadowski2018rssi,liu2007survey}.  To tackle these challenges, we develop the RSSI augmentation approach that learns the signal characteristic from the actual RSSI data using Conditional Generative Adversarial Network (ConGAN) and propose the transfer learning framework that allows RSSI sharing across different sources to benefit the indoor localisation performance of a target house.

A ConGAN is a generative neural network that generates synthetic data conditioned on specific input variables, resulting in realistic, context-specific data. The ConGANs guide the generation process using conditional information making the synthetic data more aligned with real-world patterns. This makes ConGANs particularly effective for tasks like RSSI augmentation, where they can generate diverse and realistic RSSI signals for each area of the house, enhancing the accuracy of indoor localisation models when real-world data is limited. However, sufficient among of training data is required to achieve a robust ConGAN model.

To address the challenge of insufficient training data, we apply transfer learning to the RSSI augmentation process. Instead of directly transferring RSSI data between houses, we pre-train ConGAN on a broader task, learning general RSSI characteristics from multiple houses. The pre-trained model is then fine-tuned for a target house, allowing it to generate room-specific RSSI signals while leveraging the general knowledge gained during pre-training. This approach is the first to demonstrate the transferability of RSSI data across houses with different experimental protocols, significantly improving indoor localisation performance.

The main research question we address is whether ConGAN can generate realistic RSSI data that enhances indoor localisation performance, using transfer learning to share generic RSSI information across different houses or datasets. By combining ConGAN with transfer learning, we improve the quality of the generated RSSI, allowing for better localisation in areas such as stairs and outdoor spaces, where real-world data is often limited.

Our approach outperforms traditional ConGAN methods by accelerating model convergence and enhancing performance in challenging areas. We assess this through comparisons between generated and actual RSSI data, as well as by evaluating overall F1 scores for indoor localisation performance. The main contributions of this paper are the following:
\begin{itemize}
  \item We develop a ConGAN for RSSI generation that learns room-conditioned RSSI data from a target house's RSSI signals, enabling the generation of room-specific RSSI data that significantly outperforms existing RSSI augmentation methods in term of F1 score when performing room-level localisation.
  \item We are the first to show that RSSI data can be effectively shared across diverse houses, despite variations in experimental protocols, leading to improved indoor localisation performance. We demonstrate this by proposing a transfer learning framework that uses generic RSSI information from a larger source during the RSSI augmentation pre-training process then refines the model using RSSI data from the target house during the transfer learning process. Our framework improves the generated RSSI quality indicated by the F1 score enhancement.
  \item   Building on our previous two contributions, we show that our proposed method significantly improves indoor localisation accuracy, particularly in challenging areas such as stairs and outdoor spaces, even when the training data for these areas is very limited. By initially equipping the model with the inherent characteristics of RSSI from a larger source, it can later be adapted to more unique RSSI signals through a transfer learning process, requiring only a small amount of the target RSSI data.

\end{itemize}

\section{Related Work}

Data augmentation serves as an effective strategy to enhance classifier performance across diverse applications \cite{iwana2021empirical}. 
In the domain of RSSI signal processing, considerable efforts have been made to develop data augmentation techniques by manually modifying the existing RSSI data through techniques such as the addition of noise \cite{rizk2019effectiveness,kjellson2021accurate,Suwannaphong2022}, random signal point dropping \cite{rizk2019effectiveness,kjellson2021accurate}, sampling from signal distributions \cite{rizk2019effectiveness}, and periodic signal dropping \cite{Suwannaphong2022}. These approaches aim to replicate the inherent characteristics of RSSI signals, which are characterized by noise and susceptibility to shadowing effects.

While manual augmentation is effective, automatic methods show more promise. A recent study by Njima et al. \cite{njima2021indoor} employed GAN model to synthesize RSSI values, incorporating a data selection process to retain the most realistic generated data while eliminating outliers. Similarly, a recent work by Yean et al. \cite{yean2023extendgan+} utilized a Wasserstein GAN with Gradient Penalty (WGAN-GP) in a transfer learning framework to generate RSSI data for a indoor localisation model. These works have yielded promising results by leveraging generative models to enhance indoor localisation performance. However, their focus was on WiFi RSSI in large buildings, unlike our BLE-focused study in residential areas. We design a ConGAN for BLE-based RSSI, catering to healthcare needs by emphasizing room-level information. This tailored approach prioritizes healthcare context over mere location pinpointing.

Using RSSI for indoor localisation has a limitation in sharing data across residences, hindering the generative model's knowledge of RSSI signals. Transfer learning, proven effective in various fields such as language processing, computer vision, and classification tasks, offers a solution \cite{weiss2016survey}. Prior research (Yean et al., \cite{yean2023extendgan+}) applied transfer learning with WGAN-GP for WiFi-based indoor localisation, yet only within a building. They transferred knowledge from data-rich to data-sparse zones. In contrast, our approach broadens this, transferring room-level RSSI information between residences in the same dataset and across slightly varied datasets. This empowers the generative model with generalized knowledge applicable to diverse residences.

Our study introduces the ability to share RSSI data across different houses and datasets, despite variations in layouts and room numbers—an area not previously explored. Directly transferring RSSI data is impractical due to the house-specific nature of signal characteristics. To address this, we adopt a transfer learning approach. Rather than transferring raw data, we train a ConGAN model on RSSI from multiple houses, equipping it with a broader understanding of signal patterns. Fine-tuning this model enables it to generate realistic, house-specific data, enhancing indoor localisation performance. This flexible and robust approach distinguishes our work from previous studies.

\section{Dataset}
\subsection{Annotated Residential Dataset for indoor localisation}\label{target dataset}
We use a dataset of annotated RSSI data for indoor localisation, collected by Byrne et al. \cite{byrne2018residential} in a set of representative residential homes. They used wrist-worn accelerometers sending data over BLE to Raspberry Pi-based access points (APs). The wearable transmitted data at 5Hz, and each AP recorded RSSI from the person wearing the device.

The dataset includes data from four residential homes: house A (one-bedroom apartment), B and D (two-bedroom, two-floor houses), and C (largest, three-bedroom, two-floor house). APs were strategically placed for room-level indoor localisation, with each room containing at least one AP and larger rooms equipped with multiple APs. Byrne et al.\cite{byrne2018residential} offer more details on AP locations and house layouts. 
Houses B, C, and D had 11 APs each, while house A had 8 due to its smaller size. In our study, we exclude house A due to ConGAN architecture limitations. House A's fewer APs create challenges for transferring RSSI data across houses. While adding fake APs to house A is an option, it complicates things. Data imbalance in house A—20 samples in the minority class versus up to 1271 in the majority—further complicates model evaluation. Hence, we exclude house A for simplicity in our experiments.

The dataset was created using an automated system. Binary image tags were placed on the floor at one-meter intervals. A chest-strapped camera captured images as the participant moved, generating location labels. Subsequently, a visual inspection of the video files was conducted to assess the quality of the annotations, and necessary cleaning procedures were performed to ensure accuracy. This method provided detailed training data but was time-consuming due to its complexity.

The dataset consists of two sections: fingerprint and free living. The fingerprint data is from a controlled experiment where the participant is scripted to visit every area of the house. The free-living data captures the participant's regular activities in the residence. Houses B, C, and D have scripted fingerprint measurements of 117.8, 82.0, and 71.0 minutes. The corresponding unscripted living recordings are 47.2, 237.0, and 178.4 minutes. 

\begin{figure*}
    \centering
    \includegraphics[width=0.9\linewidth]{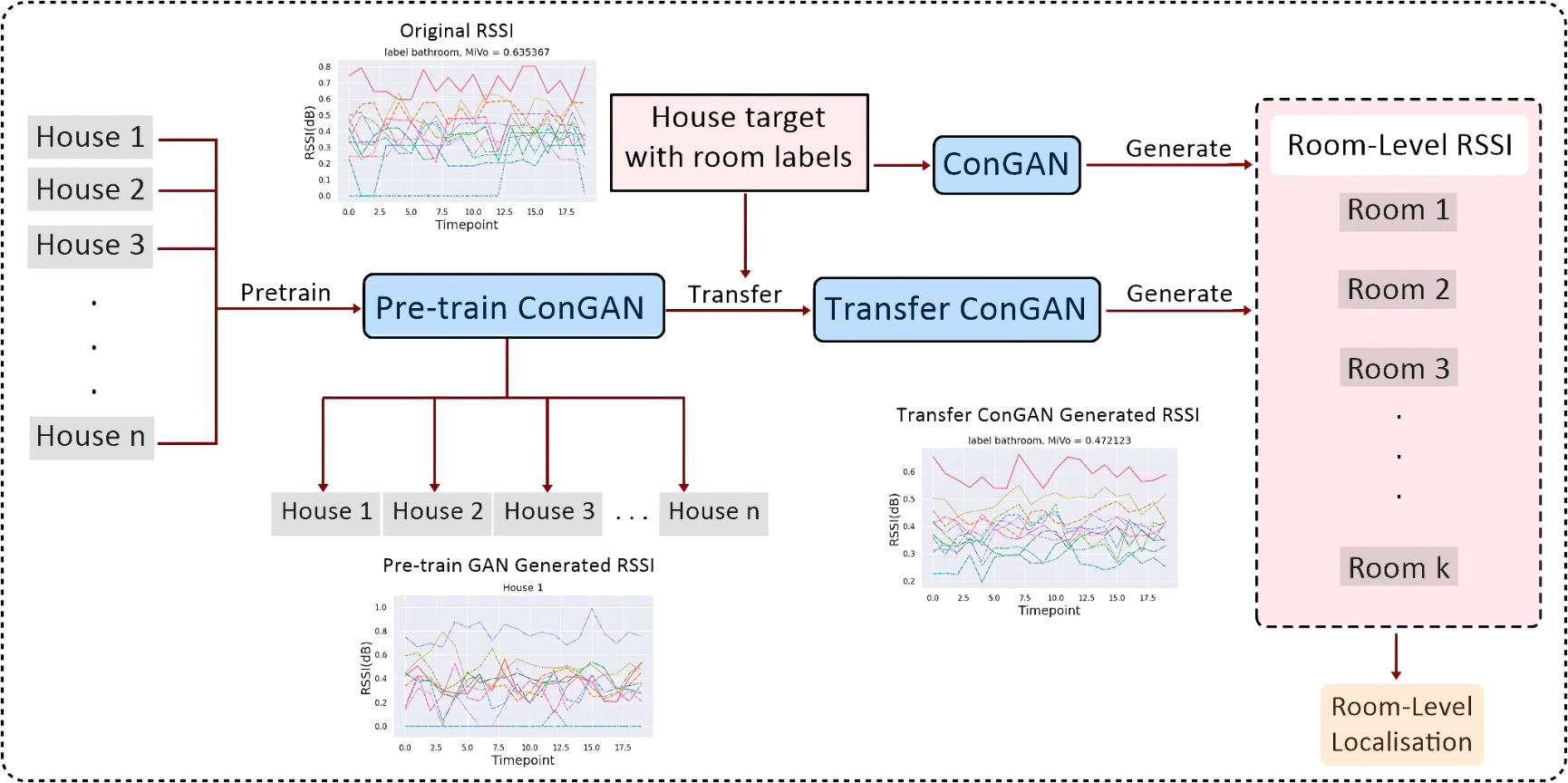}
    \caption{Transfer learning framework for RSSI augmentation. 
    }
    \label{fig:transfer_diagram}
\end{figure*}

\subsection{SPHERE Project Dataset for Transfer Learning}
We use transfer learning to enhance indoor localisation by integrating RSSI data from a larger source. We employ the SPHERE project \cite{woznowski2017sphere}, which utilises various sensors including a BLE wrist-worn accelerometer capturing RSSI. This data choice is motivated by its similarity to our target house's setup.

The SPHERE project includes data from multiple homes collected in a free-living manner. While lacking location annotations, it provides benefits like diverse RSSI data from various house types (up to 100 houses) and extensive 24/7 data for up to a year. Utilizing this abundant RSSI data from different sources has the potential to greatly improve indoor localisation performance if handled effectively. To reduce computational load, We use three SPHERE homes and extract two complete days of RSSI from each home.

Incorporating SPHERE project's RSSI data into a new house is complex due to disparate data collection protocols, sensor counts, rooms, and layouts. Differences include building types and AP numbers/types (11 Raspberry Pi APs in target houses vs. SPHERE's 9 APs, a mix of Raspberry Pi B+, NUC Intel, and NUC Intel in video cameras). Another distinction arises from data collection processes as target house data, as described in Section \ref{target dataset}, was scripted and the free-living data are not entirely natural, due to the chest-strapped camera, while SPHERE was completely free-living setup, capturing different RSSI behaviours in varied environments. However, our method demonstrates that transfer learning can indeed overcome these challenges and efficiently repurpose existing RSSI data despite the notable differences in data and environments.

\section{Methodology}

\subsection{Data preprocessing}
To handle missing data, absent values are filled forward with the latest data point for a maximum of one second, assuming RSSI values tend to remain somewhat consistent within that timeframe, with any gaps likely due to dropped packets. Additional missing values are set to -120, which isn't attainable via regular operation. Data normalization is performed using min-max normalization, and we segment the data into 4s windows with 50\% overlap.

\subsection{RSSI Augmentation Methods}
\subsubsection{Conditional GAN (ConGAN)}

The standard GAN approach involves two opposing neural networks: the generator and discriminator. The generator learns from real data to create artificial data, while the discriminator differentiates between real and artificial data \cite{goodfellow2020generative}. Their interaction results in the generator producing realistic data indistinguishable from genuine data. However, GANs have limitations for room-level indoor localisation, as they can't generate room-specific RSSI data independently. Instead, we use a ConGAN \cite{mirza2014conditional} by introducing room labels alongside RSSI data. Room labels guide the model to create room-specific RSSI data, addressing variations across rooms. We employ ConGAN to augment RSSI for each room, enhancing training data for room indoor localisation. Additionally, we use Wasserstein loss and gradient norm penalty to train ConGAN \cite{gulrajani2017improved}. Wasserstein loss calculates Wasserstein distance between real and generated distributions, aligning generated RSSI with real signals in each room. A gradient penalty (GP) ensures training stability, preventing overpowered discriminators. This stabilizes ConGAN's training, vital for accurate RSSI generation.

\begin{figure*}[h]
    \centering
    \includegraphics[width=0.95\linewidth]{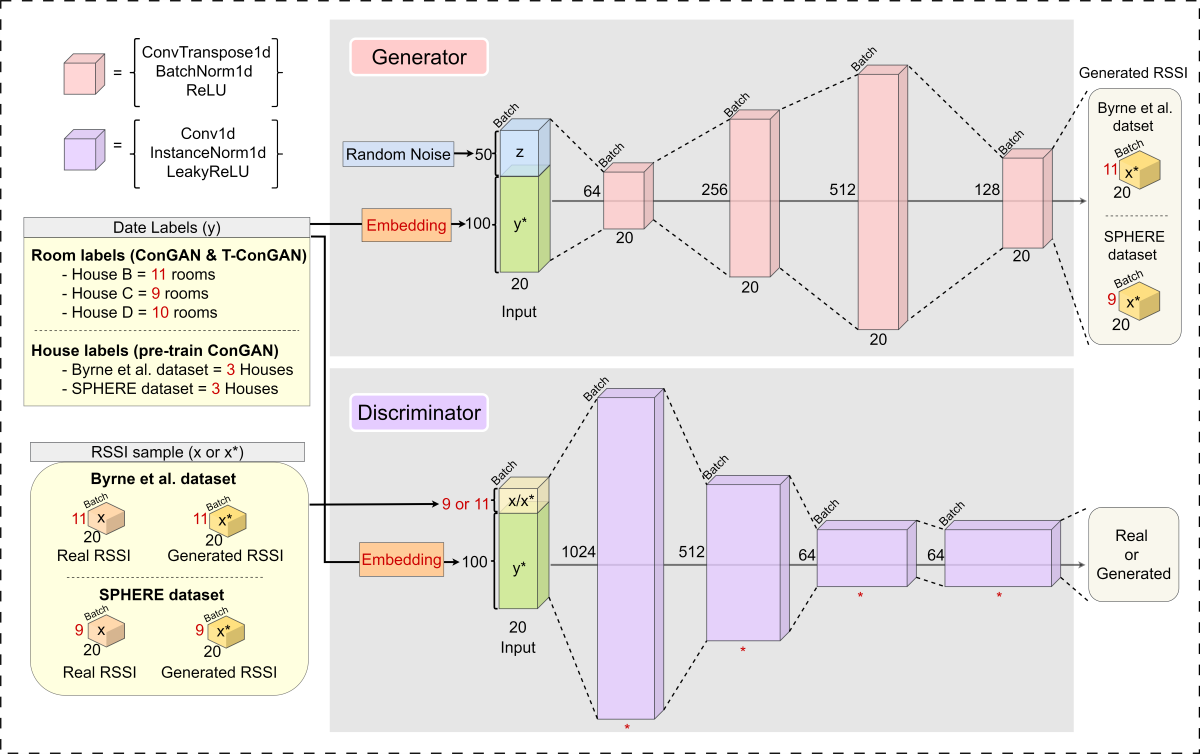}
    \caption{ConGAN architectures for the generator and discriminator. The red text highlights the architectural differences between the ConGAN during the pre-training and transfer processes. All Conv1d and ConvTranspose1d layers use a kernel size of 5, except for the final layer of the discriminator, which uses a kernel size equal to the width of the previous layer's output to produce a single value, indicating whether the RSSI input is real or generated. The red asterisks represent the dynamic width of the Conv1d output, which varies based on the input shape, as both the kernel size and output dimension are fixed.}
    \label{fig:congan}
\end{figure*}

In this study, ConGAN is implemented using Convolutional Neural Networks (CNNs). The generator includes transposed convolution layers with Batch Normalization and ReLU activation. The discriminator uses convolution layers with instance normalization and LeakyReLU activation.
Hyperparameter optimization is done via Bayesian search. Final architectures are 1D transposed convolutions [64-256-512-128] for the generator and 1D convolutions [1024-512-64-64] for the discriminator, GP of 10, batch size of 48, learning rate of 0.002077 and critic iteration of 10, found via bayesian hyperparameter optimization. The ConGAN architecture is shown in Figure \ref{fig:congan}.

\subsubsection{Transfer learning with ConGAN (T-ConGAN)}

The limitation of sharing RSSI data across different houses arises from RSSI signals being closely tied to a house's specific layout and structures. Unlike generic semantic information in images, sharing RSSI data across houses is often impractical. Sharing such data would simplify training efforts in new houses for in-home monitoring, benefiting larger deployments. ConGAN-based models are restricted to using only house-scale RSSI data, limiting generality and facing issues such as overfitting and mode collapse. Moreover, training samples may lack variability due to scripted data collection. Thus, ConGANs need greater generality to create diverse samples. Leveraging existing RSSI data can help scale up and enhance ConGANs by generating diverse artificial RSSI samples

We incorporate transfer learning to enhance the performance of the ConGAN augmentation models. The fundamental concept involves pre-training a ConGAN model using RSSI data from various houses, followed by transferring the model to each individual house, as illustrated in Figure \ref{fig:transfer_diagram}. 

The primary task is to boost ConGAN augmentation models using transfer learning. Initially, the ConGAN is pre-trained using RSSI data from multiple houses to generate RSSI data for these various houses without room labels. This pre-training does not include room labels but equips models with RSSI signal knowledge. While these pre-trained models can generate RSSI data, it lacks room specificity. The aim is to enhance ConGAN models' training data and enable RSSI data generalization from diverse sources. During transfer, the model adapts to each house, generating precise room-level RSSI data. We not only explore transfer learning between houses with the same experimental protocol but also assess it when transferring RSSI data collected under various protocols, devices, and setups.

\textbf{Transferring within the same experimental protocol}. We pre-train ConGAN using combined fingerprint (radio map) data from house B, C, and D, treating each house as a separate class, so the pre-train ConGAN generates three classes of RSSI which are RSSI from house B, C and D. Then we transfer this pre-trained ConGAN to a target house (B, C, or D) individually, to learn room-level RSSI, resulting in T-ConGAN model for each house. During transfer, we obtain the weights from pre-trained ConGAN and adapt the model's label embedding layer to match the number of rooms in the target house. Finally, we fine-tune all model weights for the specific target house.

The architectures of the pre-trained ConGAN and T-ConGAN are nearly identical, with the only difference being the label embedding layer, as shown in Figure \ref{fig:congan}. The label embedding layer converts label information into an embedded space, which is then concatenated with latent noise to form the generator’s input, or with an RSSI sample to form the discriminator’s input. This embedded label controls the generator to produces RSSI data corresponding to the provided label.

In the pre-trained ConGAN model, the label embedding layer generates 3 different embeddings corresponding to the three houses (B, C, and D). In the T-ConGAN model, the label embedding layer is adjusted to generate 11, 9, or 10 different embeddings for house B, C, and D, respectively, depending on the target house.

The PyTorch code for this embedding layer is nn.Embedding(num\_classes, embed\_size * input\_width). We use an embed\_size of 100, determined through Bayesian search, and an input\_width of 20 timestamps, which corresponds to 4 seconds of RSSI data. Other than this, both architectures are the same.

\textbf{Transferring across experimental protocols}. We pre-train ConGAN using six days of RSSI data from three SPHERE project houses. The label during this pre-training represents the specific SPHERE house, so this pre-train model generates three classes of RSSI which are RSSI from SPHERE house 1, 2 and 3. We then transfer this pre-trained model individually to house B, C, and D as new target houses to learn room-level RSSI, creating T-ConGAN-SPHERE for each target house. We adapt the model to different experimental settings by modifying relevant layers which are generator's output, discriminator's input, and label embedding layers to match the target house's APs and classes. Finally, we fine-tune the T-ConGAN-SPHERE model to account for variations in APs and classes specific to each target house (B, C, or D).

Since the number of APs differs between the SPHERE dataset (9 APs) and the Byrne et al. dataset (11 APs), the architectures of the pre-trained ConGAN with SPHERE data and T-ConGAN from SPHERE data to the target houses (B, C and D) differ more significantly compared to those within the same dataset. In addition to the difference in the number of classes that define the label embedding layer, both the size of the discriminator's input and the output size of the generator's final layer needed adjustment, as illustrated in Figure \ref{fig:congan}.

During pre-training with the SPHERE data, the label embedding layer generates 3 embeddings corresponding to the three SPHERE houses. For the T-ConGAN-SPHERE model, this is adjusted to the number of rooms in houses B, C, and D. Additionally, during the transfer process, the size of the discriminator’s input changes depending on the number of APs, as it is based on the concatenation of the RSSI sample (change from 9 APs to 11 APs) and the embedded label. The generator’s final layer is also modified to output RSSI data for 11 APs instead of 9. These changes accommodate the differences in the number of APs and classes for each task.

\subsubsection{Other Augmentations}
We used these augmentation approaches to evaluate our RSSI augmentation framework
    \begin{itemize}
        \item \textbf{SMOTE}: SMOTE (Synthetic Minority Oversampling Technique) serves as a robust baseline. SMOTE operates in feature space, creating new instances by interpolating between closely located real samples.

        \item \textbf{Domain Expert Based  Augmentations}: As a comparative augmentation technique, we explore a simple method based on domain expertise to manually mimic the nature of RSSI without changing location information. In line with the scheme from \cite{Suwannaphong2022}, we introduce noise and randomly drop APs.

        \item \textbf{Random Oversampling}: We explore random duplication of RSSI samples as an alternative method to increase the training data. This approach demonstrated competitive performance when compared to more intricate techniques \cite{batista2004study}.
    \end{itemize}

\subsection{Evaluations}
\subsubsection{Unsupervised Quality Performance}
We employ the MiVo (Mean of incoming Variance of outgoing) metric \cite{arnout2021evaluation} to assess the similarity and alignment between real and generated data. MiVo calculates both the mean and variance of the minimum pair-wise distances between generated and original samples, capturing the average and variation in distance between matched pairs of data. This metric provides a comprehensive measure of data quality and diversity, effectively addressing issues like mode collapse and overfitting \cite{arnout2021evaluation}.

\subsubsection{Classification Performance}
We evaluate the effectiveness of our proposed in improving indoor localisation performance by employing the Random Forest (RF) classifier, known for its promising performance in previous indoor localisation studies 
\cite{khokhar2021machine,jain2021low,hilal2021dataloc+}. RFs have been shown to outperform SoTA deep nets at this task \cite{jovan2023multimodal}. We optimise the RF by using a 3-fold cross-validation and perform a grid search to identify the best hyperparameters. We choose macro F1 score to comprehensively assess the model's precision and recall across various rooms, as room sizes and usage patterns can vary significantly, especially in healthcare applications where locations like stairs may be crucial despite their infrequent use.

\subsection{Experimental Setup}
The augmentation techniques were solely applied to the fingerprint data, while the free-living data was kept separate to serve as an independent evaluation. Each experiment was repeated ten times to ensure robustness and reliability, enabling the calculation of average F1 scores across all iterations. 

\textbf{Baseline Experiment}. We train classifier with fingerprint data and test on free-living data to observe the baseline performance when using only real RSSI data.

\textbf{Weighted Classifier Baseline Experiment}. In an imbalanced dataset, the minority class has a limited influence on the loss function. We introduce class weights to the baseline experiment by calculating them based on class frequencies. 

\textbf{Data Augmentation Experiments}. For each house, we augment the fingerprint data by generating up to 1000 samples for each room class. 
We employ six approaches to generate artificial RSSI:
1) Random Oversampling
2) SMOTE
3) Domain Expert Augmentations
4) ConGAN
5) T-ConGAN
and 6) T-ConGAN-SPHERE
\footnote{The code for this paper is available here:\\ \myurl.}

\begin{figure*}[h]
    \centering
    \begin{subfigure}[b]{0.3\textwidth}
         \centering
         \includegraphics[width=\textwidth, scale=0.4]{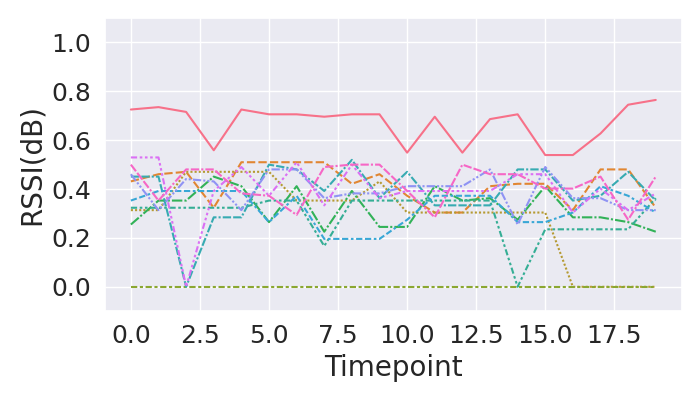}
         \caption{Original}
         \label{fig:b_bath_ori}
     \end{subfigure}

    \begin{subfigure}[b]{0.3\textwidth}
         \centering
         \includegraphics[width=\textwidth, scale=0.4]{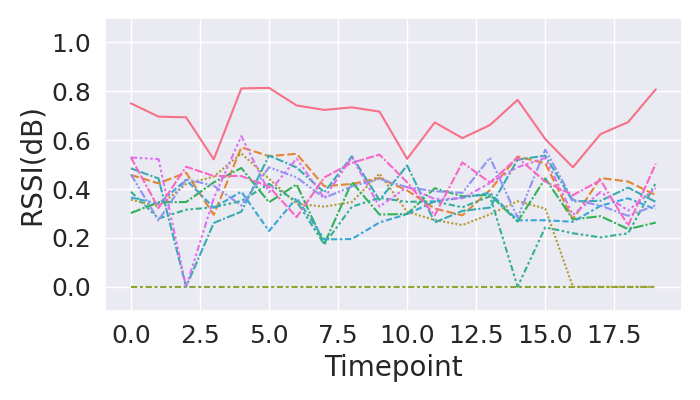}
         \caption{Domain-Expert (Noise)}
         \label{fig:b_bath_noise}
     \end{subfigure} 
     \begin{subfigure}[b]{0.3\textwidth}
         \centering
         \includegraphics[width=\textwidth, scale=0.4]{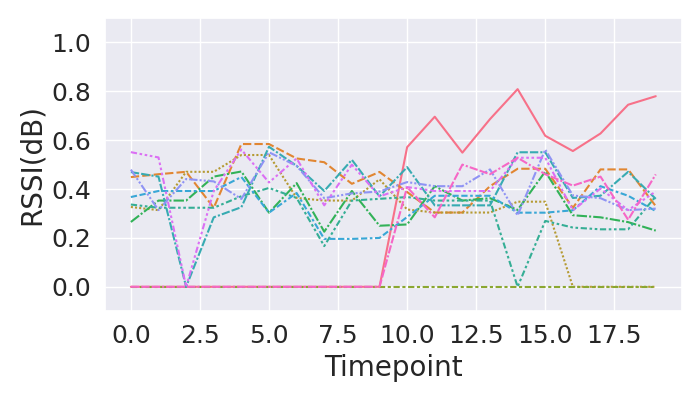}
         \caption{Domain-Expert (Drop)}
         \label{fig:b_bath_drop}
     \end{subfigure} 
     \begin{subfigure}[b]{0.3\textwidth}
         \centering
         \includegraphics[width=\textwidth, scale=0.4]{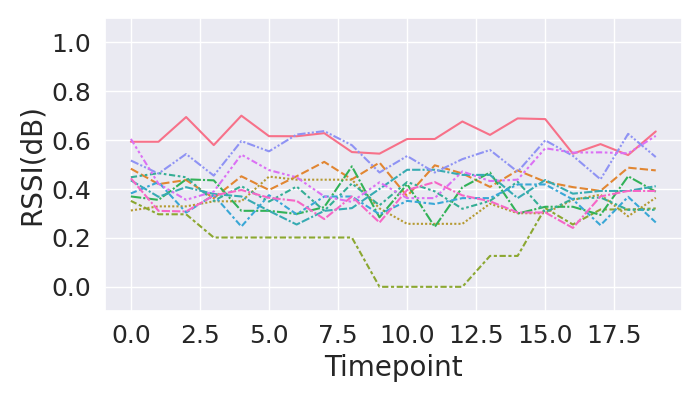}
         \caption{SMOTE}
         \label{fig:b_bath_smote}
     \end{subfigure} 
      \begin{subfigure}[b]{0.3\textwidth}
         \centering
         \includegraphics[width=\textwidth, scale=0.4]{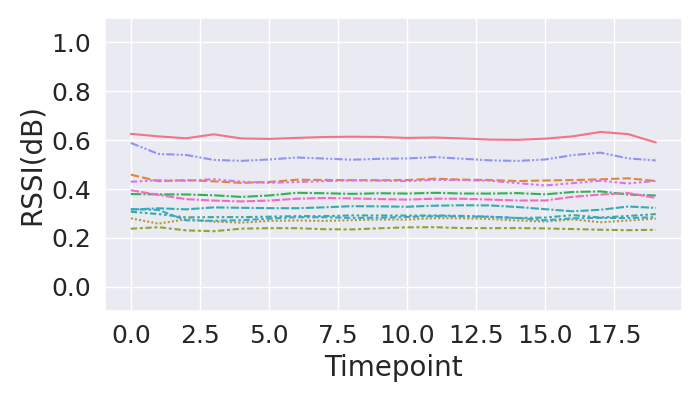}
         \caption{ConGAN}
         \label{fig:b_bath_congan}
     \end{subfigure} 
     \begin{subfigure}[b]{0.3\textwidth}
         \centering
         \includegraphics[width=\textwidth, scale=0.4]{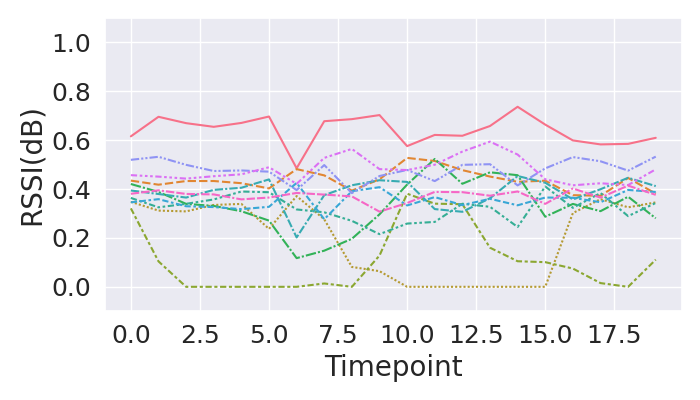}
         \caption{T-ConGAN}
         \label{fig:b_bath_tcongan}
     \end{subfigure} 
     \begin{subfigure}[b]{0.3\textwidth}
         \centering
         \includegraphics[width=\textwidth, scale=0.4]{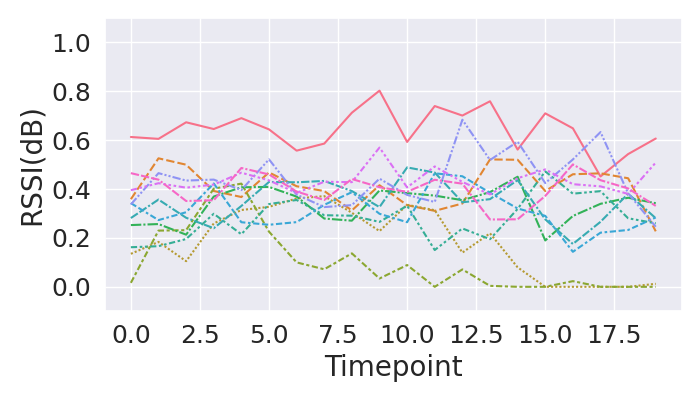}
         \caption{T-ConGAN-SPHERE}
         \label{fig:b_bath_tcongan_sphere}
     \end{subfigure} 
 
    \caption{Line plots of the actual RSSI (a) compared with augmented RSSI using different methods: (b) Domain-Expert with noise adding, (c) Domain-Expert with periodic dropping, (d) SMOTE, (e) ConGAN, (f) T-ConGAN, and (g) T-ConGAN-SPHERE. The plots consist of RSSI signals from 11 APs in residence B. Each colour represents a different AP, and we compare some example RSSI from Bathroom in a 4s window.
    }
    \label{fig:lineplot_rssi}
\end{figure*}

\begin{table}[h]
    \small
    \begin{center}
    \begin{tabular}{|l c c c|} 
     \hline
     Experiments & House B & House C & House D \\ [0.5ex] 
     \hline
     Domain-expert  & 2.0998 & 2.1387 & 2.1923 \\
     SMOTE          & \textbf{0.1375} & \textbf{0.1181} & \textbf{0.0864} \\
     ConGAN    & 0.4703          & 0.4132          & 0.4751 \\ 
     T-ConGAN    & \emph{0.4261}   & \emph{0.4077}   & \emph{0.4274} \\ 
    T-ConGAN-SPHERE  & 0.4401   & 0.4251    & 0.4351 \\ 
     \hline
    \end{tabular}
    \caption{Average MiVo across all rooms, with SMOTE lowest due to its interpolation closely matching the original RSSI, offering limited new information for localisation.}
    \label{tab:avg_mivo} 
    \end{center}
\end{table}

\section{Results and Discussions}
In this section we will present the quality of generated RSSI by comparing the distance between the generated and original RSSI data. We will also discuss the performance of the augmentation approaches for improving room-level indoor localisation.

\subsection{Unsupervised Quality Assessment}
Some methods produce RSSI data that appear very similar to real data, making it difficult to assess the quality of synthetic data through visual inspection alone, as shown in Figure \ref{fig:lineplot_rssi}. This highlights the need for a more robust metric, such as MiVo, which quantitatively measures the alignment and similarity between real and generated RSSI signals. By using MiVo, it becomes easier to compare and assess the quality of RSSI generated from different methods more objectively.

The analysis consistently shows that SMOTE yields the lowest MiVo values, as presented in Table \ref{tab:avg_mivo}. SMOTE-generated RSSI data closely resembles the original data with minimal variance, suggesting limited additional value for improving indoor localisation, as demonstrated in the next section. In contrast, T-ConGAN produces the second lowest MiVo values, higher than SMOTE but notably lower than random noise data. This indicates that T-ConGAN preserves essential location-related information while introducing valuable variance, potentially capturing RSSI variations across different parts of the room and enhancing data richness.  
However, it's essential to note that the optimal MiVo range depends on specific data characteristics, such as homes and experimental protocols. Therefore, practical evaluations like classification performance should complement MiVo for a comprehensive assessment.

\begin{table*}
\small
\caption{The macro F1 scores obtained from  RF classifier averaged from 10 runs, showcasing the superior performance achieved by transfer-based augmentation methods.}
\label{rf-f1-results}
\begin{center}
\begin{tabular}{l c c c} 
 \hline
 Experiments & House B & House C & House D \\ [0.5ex] 
 \hline
 Baseline (no augmentation)  & 71.67$\pm$1.02   & 68.96$\pm$0.70   & 71.07$\pm$0.44 \\ 
 Weighted Classifier  & 71.39$\pm$1.01   & 68.62$\pm$1.27   & 70.02$\pm$0.28 \\
 Random Oversampling  & 73.12$\pm$0.67   & 73.02$\pm$0.67   & 72.15$\pm$0.59 \\
 SMOTE       & \emph{76.77}$\pm$0.81      & 77.91$\pm$0.51          & 74.67$\pm$0.80 \\
 Domain-Expert      & 74.72$\pm$0.86   & 79.14$\pm$0.99   & 75.68$\pm$0.61 \\
 ConGAN     & 66.03$\pm$0.85      &\emph{79.60}$\pm$0.73    & 69.28$\pm$0.39 \\ 
 T-ConGAN      & 74.48$\pm$0.79      & \textbf{81.16}$\pm$0.55 & \textbf{79.17}$\pm$0.59 \\ 
 T-ConGAN-SPHERE & \textbf{77.21}$\pm$0.69   &79.42$\pm$0.57 & \emph{78.57}$\pm$0.67 \\
 \hline
\end{tabular}

\end{center}
\end{table*}

\subsection{Supervised Quality Assessment}

Our supervised quality assessment utilises ground truth labels to measure how well each approach improves indoor localisation performance, and we will show how transfer learning approaches lead to the largest improvements.

In general, T-ConGAN shows the highest performance among all augmentation approaches, except in the case of house B, where T-ConGAN-SPHERE performs the best. Table \ref{rf-f1-results} illustrates the positive impact of data augmentation on indoor localisation performance. Most augmentation methods, except ConGAN for house B and D, lead to an increase in the macro F1 score compared to the baseline experiment. However, the transfer-based augmentation method consistently outperforms other approaches, underscoring the significance of transfer learning from a larger source of existing RSSI data. 

Although house B and house D are similar, as both are two-bedroom, two-floor houses, T-ConGAN-SPHERE and T-ConGAN perform better in house D than in house B. The reason for this is because of the complexity of house B's layout, particularly the stairs, which are split into two levels due to the bathroom being situated on a landing between the first and second floors. The dataset labels these levels separately as upper stairs and lower stairs, but no APs are placed directly on the stairs to support accurate localisation. This makes it difficult to distinguish between the two stair locations, significantly lowering the macro F1 score, even with data augmentation. The classification accuracy for the lower and upper stairs without augmentation is around 3\% and 30\%, respectively, as the two classes often confuse the model. However, T-ConGAN-SPHERE outperforms T-ConGAN in house B, possibly also due to the unique characteristics of house B’s layout. The added variation in RSSI data from the SPHERE dataset may provide better differentiation between the stair locations than data from within the same experimental protocol, resulting in improved RSSI generation and higher localisation performance.

Various methods were explored to address data limitations and improve indoor localisation performance. The weighted classifier baseline, targeting class imbalance, showed little impact on the RF classifier's performance, implying that data imbalance does not significantly affect it. Random oversampling, designed to expand limited training data, demonstrated moderate performance improvements. SMOTE, enhancing both data quantity and quality, outperformed random oversampling, emphasizing the value of high-quality augmented data. Domain-expert augmentation, manually tuning approach, impressively improved performance but was not consistently superior to SMOTE. ConGAN, an automated approach, showed mixed results across houses, with some improvements and decreases in performance. T-ConGAN leveraged pre-trained models and improved performance consistently across houses. T-ConGAN-SPHERE extended this by using data from different experimental protocols, achieving favorable performance without additional target house data collection. These results indicate that our transfer-based augmentation method outperform all other augmentation approaches.

\subsection{Why Transfer Learning Improves indoor localisation Performance}
Transfer learning significantly boosts indoor localisation performance, particularly for minority classes, which are characterized by a smaller number of labelled samples compared to other classes, resulting in poorer indoor localisation performance for these classes. The performance of minority classes typically struggles, often falling below 40\% and occasionally as low as 4\%. This substantial discrepancy significantly hampers overall indoor localisation performance. The results in Table \ref{min-class} show the best improvement being a remarkable 51\% increase over the baseline in house C for outside class, using just 3 minutes of scripted fingerprint data along with generated RSSI. The other notable improvements are 32\%, 30\% and 40.9\% in stairs classes of house B, C and D where these classes only has 5.47, 8.07 and 6.07 mins of training data. These results indicate that our transferred GANs can derive valuable insights from RSSI data collected from various sources, even beyond the target area, and generate precise location-specific RSSI data, outperforming models trained with limited samples.

\begin{table}
\small
\caption{ Class accuracy improvement achieved by the best model of each house from Table \ref{rf-f1-results}, highlighted that our transfer-based method can improve the class accuracy by up to 51 \% even with the training data as little as 3 mins. 
}
\label{min-class}
\begin{center}
\setlength{\tabcolsep}{2pt} 
\begin{tabular}{|c|l|c|c|} 
 \hline
 House & Classes & Training Data &  Improvement\\ [0.5ex] 
 \hline
 B    & Lower stairs   & 5.47 mins   & 32.0\% \\ 
            & Upper stairs   & 8.60 mins   & 21.0\% \\ 
            \hline
 C & Outside & 3.00 mins & 51.0\% \\
         & Stairs  & 8.07 mins & 30.0\% \\
 \hline
 D & Outside & 3.47 mins & 29.0\% \\
         & Stairs  & 6.07 mins & 40.9\% \\
 \hline
\end{tabular}
\end{center}
\end{table}

The improved performance in challenging areas such as stairs and outdoor spaces is highly valuable in healthcare applications. These areas are prone to emergencies and can be difficult for elderly patients to navigate. This finding enhances the monitoring system's ability to support vulnerable individuals in these areas, ultimately improving their quality of life.

\subsection{Limitations}
Our system is not designed to address emergency scenarios. Instead, the primary goal of our framework is to develop a low-cost and scalable indoor localisation system, which can be easily deployed by individual patients with minimal effort required for data collection and system setup. This makes it practical for large-scale use in health monitoring applications such as early diagnosis, post-treatment monitoring, or personalised care. 

In these non-critical applications, the impact of false positives and negatives is relatively minor. However, for emergency situations that require life-saving responses, specialised systems with extremely high accuracy and sensitivity would be needed. Such systems may involve additional hardware capable of notifying carers in real-time with precise location information, which would significantly raise the cost. Given the focus on affordability and scalability in our current work, developing high-criticality systems for emergency scenarios falls outside the scope of this study.

While our framework effectively uses BLE data, its compatibility with other signal types such as WiFi or UWB remains uncertain and requires further exploration with diverse datasets. Our work suggests reduced data collection efforts for system training; however, further investigations are needed to validate the system in real-world environments with different physical characteristics and configurations.

\section{Future Work}
There are several areas for future exploration to broaden the scope and applicability of the system. While the number of APs used in our experiments is considered realistic for achieving accurate room-level indoor localisation in homes of this size, with at least one AP per room as seen in similar studies \cite{kyritsis2016ble, garcia2021empirical, tegou2018low}, we aim to extend our work to environments with fewer APs and smaller residential areas to assess the system's robustness in more constrained settings. 

Our work serves as a validation of the transfer learning capability, that will play a key role in the future studies. We will expand the diversity of homes in future studies, including homes with thicker concrete walls and more compartmentalised layouts, as our current study is limited to typical UK homes in terms of layout, furniture style, and construction materials.

Furthermore, although the current study was conducted in actual residential homes, we recognise that a wider variety of environments, including those with more signal interference and different physical configurations, need to be explored to further validate the system. Testing in more diverse environments will provide a better understanding of how our system performs under varying real-world conditions.

In terms of data sharing and transfer learning, we have shown that RSSI data can be shared across different houses, but the variation in this study is limited to homes with similar physical characteristics. We will investigate the potential impact of irrelevant data during the pre-training phase, especially when transferring data between homes with significant physical differences. Additionally, we plan to confirm the model’s performance under variations in the number and location of APs, as the current study does not cover this aspect comprehensively.

Our framework focuses on BLE RSSI because of its low power consumption \cite{obeidat2021review}, making it ideal for in-home applications, especially for elderly individuals.  However, we have not explored other technologies like Wi-Fi or UWB, which are better suited for large-scale settings \cite{obeidat2021review}, such as large building navigation \cite{ng2019environmental} or asset tracking \cite{ahmed2020comparative}, where power consumption is less of a concern, data collection is easier, and privacy constraints are less restrictive. Future work will consider investigating these technologies to broaden the scope of this research.

Lastly, while overfitting could be an issue in houses with very different characteristics, we believe this can be mitigated by selecting homes that are both similar and different enough to provide the sufficient variation for generalisation. In future work, we will further explore this by carefully balancing the selection of houses to ensure the model's adaptability across different environments.

\section{Conclusion}

In this study, we propose a ConGAN-based augmentation method combined with our transfer learning framework to improve room-level indoor localisation in smart homes. Our approach address the problem of limited number of training data for in-home monitoring system, and we are the first work that demonstrates the ability to transfer RSSI across homes and improve indoor localisation performance. The results demonstrate that augmentation enhances indoor localisation performance, but our transfer learning framework improves the performance further. This is the state-of-the-art model for RSSI augmentation, enabling the utilisation of the large amounts of existing, unannotated free-living datasets to improve performance and scale. In particular, we show how this approach improves performance for minority classes in the dataset. For healthcare purposes, our proposed approach can reduce the patient effort in collecting training data, and improves location monitoring performance on stairs and outside the home, which are crucial areas for the patients with limited mobility.

\bibliographystyle{ieeetr}
\bibliography{ref}

\begin{thebibliography}{10}

\bibitem{poyiadzi2020detecting}
R.~Poyiadzi, W.~Yang, Y.~Ben-Shlomo, I.~Craddock, L.~Coulthard, R.~Santos-Rodriguez, J.~Selwood, and N.~Twomey, ``Detecting signatures of early-stage dementia with behavioural models derived from sensor data,'' {\em arXiv preprint arXiv:2007.03615}, 2020.

\bibitem{alarifi2016ultra}
A.~Alarifi, A.~Al-Salman, M.~Alsaleh, A.~Alnafessah, S.~Al-Hadhrami, M.~A. Al-Ammar, and H.~S. Al-Khalifa, ``Ultra wideband indoor positioning technologies: Analysis and recent advances,'' {\em Sensors}, vol.~16, no.~5, p.~707, 2016.

\bibitem{bennett2017healthcare}
J.~Bennett, O.~Rokas, and L.~Chen, ``Healthcare in the smart home: A study of past, present and future,'' {\em Sustainability}, vol.~9, no.~5, p.~840, 2017.

\bibitem{van2020indoor}
W.~Van~Woensel, P.~C. Roy, S.~S.~R. Abidi, and S.~R. Abidi, ``Indoor location identification of patients for directing virtual care: An ai approach using machine learning and knowledge-based methods,'' {\em AI in Medicine}, vol.~108, p.~101931, 2020.

\bibitem{byrne2018residential}
D.~Byrne, M.~Kozlowski, R.~Santos-Rodriguez, R.~Piechocki, and I.~Craddock, ``Residential wearable rssi and accelerometer measurements with detailed location annotations,'' {\em Scientific data}, vol.~5, no.~1, pp.~1--14, 2018.

\bibitem{McConville2021}
R.~McConville, G.~Archer, I.~Craddock, M.~Kozłowski, R.~Piechocki, J.~Pope, and R.~Santos-Rodriguez, ``Vesta: A digital health analytics platform for a smart home in a box,'' {\em Future Generation Computer Systems}, vol.~114, pp.~106 -- 119, 2021.

\bibitem{Suwannaphong2022}
T.~Suwannaphong, R.~McConville, and I.~Craddock, ``Radio signal strength indication augmentation for one-shot learning in indoor localisation,'' in {\em SmartWear '22}, p.~7–12, ACM, 2022.

\bibitem{sadowski2018rssi}
S.~Sadowski and P.~Spachos, ``Rssi-based indoor localization with the internet of things,'' {\em IEEE access}, vol.~6, pp.~30149--30161, 2018.

\bibitem{liu2007survey}
H.~Liu, H.~Darabi, P.~Banerjee, and J.~Liu, ``Survey of wireless indoor positioning techniques and systems,'' {\em IEEE Transactions on Systems, Man, and Cybernetics, Part C (Applications and Reviews)}, vol.~37, no.~6, pp.~1067--1080, 2007.

\bibitem{iwana2021empirical}
B.~K. Iwana and S.~Uchida, ``An empirical survey of data augmentation for time series classification with neural networks,'' {\em Plos one}, vol.~16, no.~7, p.~e0254841, 2021.

\bibitem{rizk2019effectiveness}
H.~Rizk, A.~Shokry, and M.~Youssef, ``Effectiveness of data augmentation in cellular-based localization using deep learning,'' in {\em WCNC}, pp.~1--6, IEEE, 2019.

\bibitem{kjellson2021accurate}
C.~Kjellson, M.~Larsson, K.~{\"A}str{\"o}m, and M.~Oskarsson, ``Accurate indoor positioning based on learned absolute and relative models,'' in {\em IPIN}, pp.~1--8, IEEE, 2021.

\bibitem{njima2021indoor}
W.~Njima, M.~Chafii, A.~Chorti, R.~M. Shubair, and H.~V. Poor, ``Indoor localization using data augmentation via selective generative adversarial networks,'' {\em IEEE Access}, vol.~9, pp.~98337--98347, 2021.

\bibitem{yean2023extendgan+}
S.~Yean, W.~Goh, B.-S. Lee, and H.~L. Oh, ``extendgan+: Transferable data augmentation framework using wgan-gp for data-driven indoor localisation model,'' {\em Sensors}, vol.~23, no.~9, p.~4402, 2023.

\bibitem{weiss2016survey}
K.~Weiss, T.~M. Khoshgoftaar, and D.~Wang, ``A survey of transfer learning,'' {\em Journal of Big data}, vol.~3, no.~1, pp.~1--40, 2016.

\bibitem{woznowski2017sphere}
P.~Woznowski, A.~Burrows, T.~Diethe, X.~Fafoutis, J.~Hall, S.~Hannuna, M.~Camplani, N.~Twomey, M.~Kozlowski, B.~Tan, {\em et~al.}, ``Sphere: A sensor platform for healthcare in a residential environment,'' {\em Designing, Developing, and Facilitating Smart Cities: Urban Design to IoT Solutions}, pp.~315--333, 2017.

\bibitem{goodfellow2020generative}
I.~Goodfellow, J.~Pouget-Abadie, M.~Mirza, B.~Xu, D.~Warde-Farley, S.~Ozair, A.~Courville, and Y.~Bengio, ``Generative adversarial networks,'' {\em Communications of the ACM}, vol.~63, no.~11, pp.~139--144, 2020.

\bibitem{mirza2014conditional}
M.~Mirza and S.~Osindero, ``Conditional generative adversarial nets,'' {\em arXiv preprint arXiv:1411.1784}, 2014.

\bibitem{gulrajani2017improved}
I.~Gulrajani, F.~Ahmed, M.~Arjovsky, V.~Dumoulin, and A.~C. Courville, ``Improved training of wasserstein gans,'' {\em Advances in neural information processing systems}, vol.~30, 2017.

\bibitem{batista2004study}
G.~E. Batista, R.~C. Prati, and M.~C. Monard, ``A study of the behavior of several methods for balancing machine learning training data,'' {\em ACM SIGKDD explorations newsletter}, vol.~6, no.~1, pp.~20--29, 2004.

\bibitem{arnout2021evaluation}
H.~Arnout, J.~Bronner, and T.~Runkler, ``Evaluation of generative adversarial networks for time series data,'' in {\em IJCNN}, pp.~1--7, IEEE, 2021.

\bibitem{khokhar2021machine}
Z.~Khokhar and M.~A. Siddiqi, ``Machine learning based indoor localization using wi-fi and smartphone,'' {\em J. Independent Stud. Res. Comput}, vol.~18, no.~1, 2021.

\bibitem{jain2021low}
C.~Jain, G.~V.~S. Sashank, S.~Markkandan, {\em et~al.}, ``Low-cost ble based indoor localization using rssi fingerprinting and machine learning,'' in {\em WiSPNET}, pp.~363--367, IEEE, 2021.

\bibitem{hilal2021dataloc+}
A.~Hilal, I.~Arai, and S.~El-Tawab, ``Dataloc+: A data augmentation technique for machine learning in room-level indoor localization,'' in {\em WCNC}, pp.~1--7, IEEE, 2021.

\bibitem{jovan2023multimodal}
F.~Jovan, C.~Morgan, R.~McConville, E.~L. Tonkin, I.~Craddock, and A.~Whone, ``Multimodal indoor localisation in parkinson's disease for detecting medication use: Observational pilot study in a free-living setting,'' in {\em Proceedings of the 29th ACM SIGKDD Conference on Knowledge Discovery and Data Mining}, pp.~4273--4283, 2023.

\bibitem{kyritsis2016ble}
A.~I. Kyritsis, P.~Kostopoulos, M.~Deriaz, and D.~Konstantas, ``A ble-based probabilistic room-level localization method,'' in {\em 2016 International Conference on Localization and GNSS (ICL-GNSS)}, pp.~1--6, IEEE, 2016.

\bibitem{garcia2021empirical}
P.~J. Garc{\'\i}a-Paterna, A.~S. Mart{\'\i}nez-Sala, and J.~C. S{\'a}nchez-Aarnoutse, ``Empirical study of a room-level localization system based on bluetooth low energy beacons,'' {\em Sensors}, vol.~21, no.~11, p.~3665, 2021.

\bibitem{tegou2018low}
T.~Tegou, I.~Kalamaras, K.~Votis, and D.~Tzovaras, ``A low-cost room-level indoor localization system with easy setup for medical applications,'' in {\em 2018 11th IFIP Wireless and Mobile Networking Conference (WMNC)}, pp.~1--7, IEEE, 2018.

\bibitem{obeidat2021review}
H.~Obeidat, W.~Shuaieb, O.~Obeidat, and R.~Abd-Alhameed, ``A review of indoor localization techniques and wireless technologies,'' {\em Wireless Personal Communications}, vol.~119, pp.~289--327, 2021.

\bibitem{ng2019environmental}
J.~K. Ng, H.~Li, V.~C. Cheng, and W.~K. Cheung, ``An environmental-adaptive wi-fi localization approach with low start-up cost for the exhibition industry,'' in {\em 2019 IEEE 25th International Conference on Embedded and Real-Time Computing Systems and Applications (RTCSA)}, pp.~1--8, IEEE, 2019.

\bibitem{ahmed2020comparative}
F.~Ahmed, M.~Phillips, S.~Phillips, and K.-Y. Kim, ``Comparative study of seamless asset location and tracking technologies,'' {\em Procedia Manufacturing}, vol.~51, pp.~1138--1145, 2020.

\end{thebibliography}

\end{document}